\documentclass[usenames,dvipsnames]{article} 
\pdfoutput=1
\usepackage{nips13submit_e,times}
\usepackage{url}
\usepackage{subfig,wrapfig,floatrow}
\usepackage{graphicx}
\usepackage[square,sort,comma,numbers]{natbib}
\usepackage{lipsum}
\usepackage{paralist}
\usepackage{amsmath,amssymb,amsthm}
\usepackage[]{units}
\usepackage[scientific-notation=true]{siunitx}
\usepackage{url}

\renewcommand{\vec}[1]{\mbox{\boldmath$#1$}}

\title{Deep learning for neuroimaging: a validation study}

\author{
Sergey M. Plis \\
The Mind Research Network\\
Albuquerque, NM 87106 \\
\texttt{s.m.plis@gmail.com} \\
\And
Devon R. Hjelm \\
University of New Mexico\\
Albuquerque, NM 87131\\
\texttt{dhjelm@mrn.org} \\
\AND
Ruslan Salakhutdinov \\
University of Toronto\\
Toronto, Ontario M5S 2E4 \\
\texttt{rsalakhu@cs.toronto.edu} \\
\And
Vince D. Calhoun\\
The Mind Research Network\\
Albuquerque, NM 87106\\
\texttt{vcalhoun@mrn.org} \\
}

\nipsfinalcopy 
\graphicspath{
{figures/}
}

\begin{document}

\maketitle

\begin{abstract}
  Deep learning  methods have  recently made  notable advances  in the
  tasks of  clas- sification and representation  learning. These tasks
  are important  for brain imaging and  neuroscience discovery, making
  the  methods  attractive for  porting  to  a neuroimager's  toolbox.
  Success of these  methods is, in part, explained  by the flexibility
  of deep learning models. However, this flexibility makes the process
  of  porting   to  new  areas  a   difficult  parameter  optimization
  problem.  In this  work  we demonstrate  our  results (and  feasible
  parameter  ranges)  in  application  of  deep  learning  methods  to
  structural and  functional brain imaging  data.  We also  describe a
  novel  constraint-based  approach  to visualizing  high  dimensional
  data. We use it to ana- lyze the effect of parameter choices on data
  transformations.  Our  results show  that deep learning  methods are
  able to  learn physiologically important representations  and detect
  latent relations in neuroimaging data.
\end{abstract}

\section{Introduction}
\label{sec:intro}
One  of  the  main  goals of  brain  imaging  and  neuroscience---and,
possibly, of  most natural  sciences---is to improve  understanding of
the investigated system  based on data.  In our case,  this amounts to
inference of descriptive features of brain structure and function from
non-invasive measurements.   Brain imaging field  has come a  long way
from anatomical maps and atlases  towards data driven feature learning
methods,  such as  seed-based correlation~\cite{biswal1995functional},
canonical   correlation   analysis~\cite{Sui2012},   and   independent
component  analysis  (ICA)~\cite{mckeown1998analysis,BellSejnowski95}.
These methods are highly successful  in revealing known brain features
with  new  details~\cite{brookes2011investigating}  (supporting  their
credibility), in  recovering features that differentiate  patients and
controls~\cite{Potluru2008}    (assisting   diagnosis    and   disease
understanding),  and starting  a  ``resting  state'' revolution  after
revealing  consistent  patters  in   data  from  uncontrolled  resting
experiments~\cite{van2010exploring,raichle16012001}.    Classification
is  often used  merely as  a correctness  checking tool,  as the  main
emphasis is  on learning about the  brain. A perfect oracle  that does
not explain its conclusions would  be useful, but mainly to facilitate
the inference of the ways the oracle draws these conclusions.

As an  oracle, deep learning  methods are breaking records  taken over
the  areas  of  speech,  signal,  image, video  and  text  mining  and
recognition by improving state of  the art classification accuracy by,
sometimes, more than 30\% where the prior decade struggled to obtain a
1-2\% improvements~\cite{krizhevsky2012,lee2012}.  What differentiates
them  from  other  classifiers,  however,  is  the  automatic  feature
learning  from  data  which  largely contributes  to  improvements  in
accuracy.   Presently, this  seems to  be the  closest solution  to an
oracle  that  reveals its  methods  ---  a  desirable tool  for  brain
imaging.

Another distinguishing  feature of deep  learning is the depth  of the
models.  Based on already acceptable feature learning results obtained
by  shallow models---currently  dominating neuroimaging  field---it is
not immediately clear what benefits would depth have.  Considering the
state of  multimodal learning, where  models are either assumed  to be
the   same   for   analyzed   modalities~\cite{moosmann2008joint}   or
cross-modal relations  are sought  at the  (shallow) level  of mixture
coefficients~\cite{liu2007parallel},  deeper  models  better  fit  the
intuitive  notion  of  cross-modality   relations,  as,  for  example,
relations  between   genetics  and  phenotypes  should   be  indirect,
happening at a deeper conceptual level.

In this  work we present  our recent  advances in application  of deep
learning  methods  to  functional and  structural  magnetic  resonance
imaging (fMRI and  sMRI). Each consists of brain volumes  but for sMRI
these are static volumes---one per subject/session,---while for fMRI a
single subject dataset is comprised  of multiple volumes capturing the
changes  during an  experimental  session.  Our  goal  is to  validate
feasibility of this application by
\begin{inparaenum}[\itshape   a\upshape)]  
\item\label{itm:rbm}  investigating  if  a   building  block  of  deep
  generative      models---a     restricted      Boltzmann     machine
  (RBM)~\cite{hinton2000}---is competitive with  ICA (a representative
  model of its class) (Section~\ref{sec:rbm});
\item\label{itm:dpth}  examining  the  effect  of the  depth  in  deep
  learning analysis of structural MRI data (Section~\ref{sec:depth}); and
\item\label{itm:predict}  determining the  value  of  the methods  for
  discovery  of latent  structure  of a  large-scale (by  neuroimaging
  standards) dataset (Section~\ref{sec:discovery}).
\end{inparaenum}
The   measure   of  feature   learning   performance   in  a   shallow
model~(\ref{itm:rbm}) is  comparable with  existing methods  and known
brain physiology.   However, this measure  cannot be used  when deeper
models are  investigated.  As  we further  demonstrate, classification
accuracy does not  provide the complete picture either. To  be able to
visualize the  effect of depth and  gain an insight into  the learning
process,  we introduce  a flexible  constraint satisfaction  embedding
method that  allows us  to control the  complexity of  the constraints
(Section~\ref{sec:dandc}).  Deliberately choosing local constraints we
are able to  reflect the transformations that the  deep belief network
(DBN)~\cite{HinSal06}  learns  and  applies   to  the  data  and  gain
additional insight.

\section{A shallow belief network for feature learning}
\label{sec:rbm}

Prior to  investigating the  benefits of  depth of  a DBN  in learning
representations from fMRI and sMRI data,  we would like to find out if
a shallow  (single hidden layer)  model--which is the  RBM---from this
family  meets   the  field's   expectations.   As  mentioned   in  the
introduction, a number  of methods are used for  feature learning from
neuroimaging  data:  most   of  them  belong  to   the  single  matrix
factorization (SMF) class.  We do a quick comparison to a small subset
of SMF methods  on simulated data; and continue with  a more extensive
comparison  against ICA  as an  approach trusted  in the  neuroimaging
field. Similarly to RBM, ICA  relies on the bipartite graph structure,
or even is  an artificial neural network with sigmoid  hidden units as
is in the  case of Infomax ICA~\cite{BellSejnowski95}  that we compare
against.  Note the difference with  RBM: ICA applies its weight matrix
to the (shorter) temporal dimension  of the data imposing independence
on the spatial  dimension while RBM applies its  weight matrix (hidden
units ``receptive fields'') to  the high dimensional spatial dimension
instead (Figure~\ref{fig:pipeline}).

\subsection{A restricted Boltzmann machine}
\label{sec:RBM}
A \emph{restricted Boltzmann  machine} (RBM) is a  Markov random field
that  models  data  distribution  parameterizing  it  with  the  Gibbs
distribution  over a  bipartite  graph between  visible $\vec{v}$  and
hidden    variables     $\vec{h}$~\cite{fischer2012introduction}:    $
p(\vec{v})  =  \sum_{\vec{h}}p(\vec{v},\vec{h}) =  \sum_{\vec{h}}  1/Z
\exp(-E(\vec{v},\vec{h})),  $   where  $Z=\sum_{\vec{v}}\sum_{\vec{h}}
e^{-E(\vec{v},\vec{h})}$  is  the  normalization term  (the  partition
function) and $E(\vec{v}, \vec{h})$ is the energy of the system.  Each
visible variable  in the case  of fMRI data  represents a voxel  of an
fMRI scan with a  real-valued and approximately Gaussian distribution.
In this case, the energy is defined as:
\begin{align}
  E(\vec{v},\vec{h}) &=  - \sum_{ij}\frac{v_j}{\sigma_j} W_{ji}  h_i -
  \sum_j \frac{(a_j-v_j)^2}{\sigma_j^2} - \sum_i b_ih_i,
\end{align}
where  $a_j$ and  $b_j$  are  biases and  $\sigma_j$  is the  standard
deviation  of  a  parabolic  containment  function  for  each  visible
variable $v_j$ centered on the bias $a_j$.  In general, the parameters
$\sigma_i$  need  to  be  learned along  with  the  other  parameters.
However, in  practice normalizing  the distribution  of each  voxel to
have   zero   mean    and   unit   variance   is    faster   and   yet
effective~\cite{nair2010rectified}.   A number  of choices  affect the
quality of interpretation of the  representations learned from fMRI by
an  RBM.  Encouraging  sparse features  via the  $L_1$-regularization:
\(\lambda  \|W\|_1\) (\(\lambda=0.1\)  gave  best  results) and  using
hyperbolic  tangent  for  hidden  units  non-linearity  are  essential
settings   that   respectively   facilitate   spatial   and   temporal
interpretation  of the  result.  The  weights were  updated using  the
truncated  Gibbs sampling  method called  contrastive divergence  (CD)
with a single  sampling step (CD-1). Further information  on RBM model
can be found in~\cite{hinton2000,hinton2006fast}.

\subsection{Synthetic data}
\label{sec:syntheticdata}
\begin{figure}[ht]%
\captionsetup[subfigure]{style=default, margin=4pt, parskip=0pt,
  hangindent=0pt, indention=0pt, singlelinecheck=true, font=footnotesize}
\vspace{-15pt}
\centering
\subfloat[Average spatial map (SM) and time course (TC) correlations
to ground truth  for RBM and SMF models (gray box).]{
  \begin{minipage}[c][0.22\textwidth]{%
      0.3\textwidth}
    \centering%
    \includegraphics[width=1\textwidth]{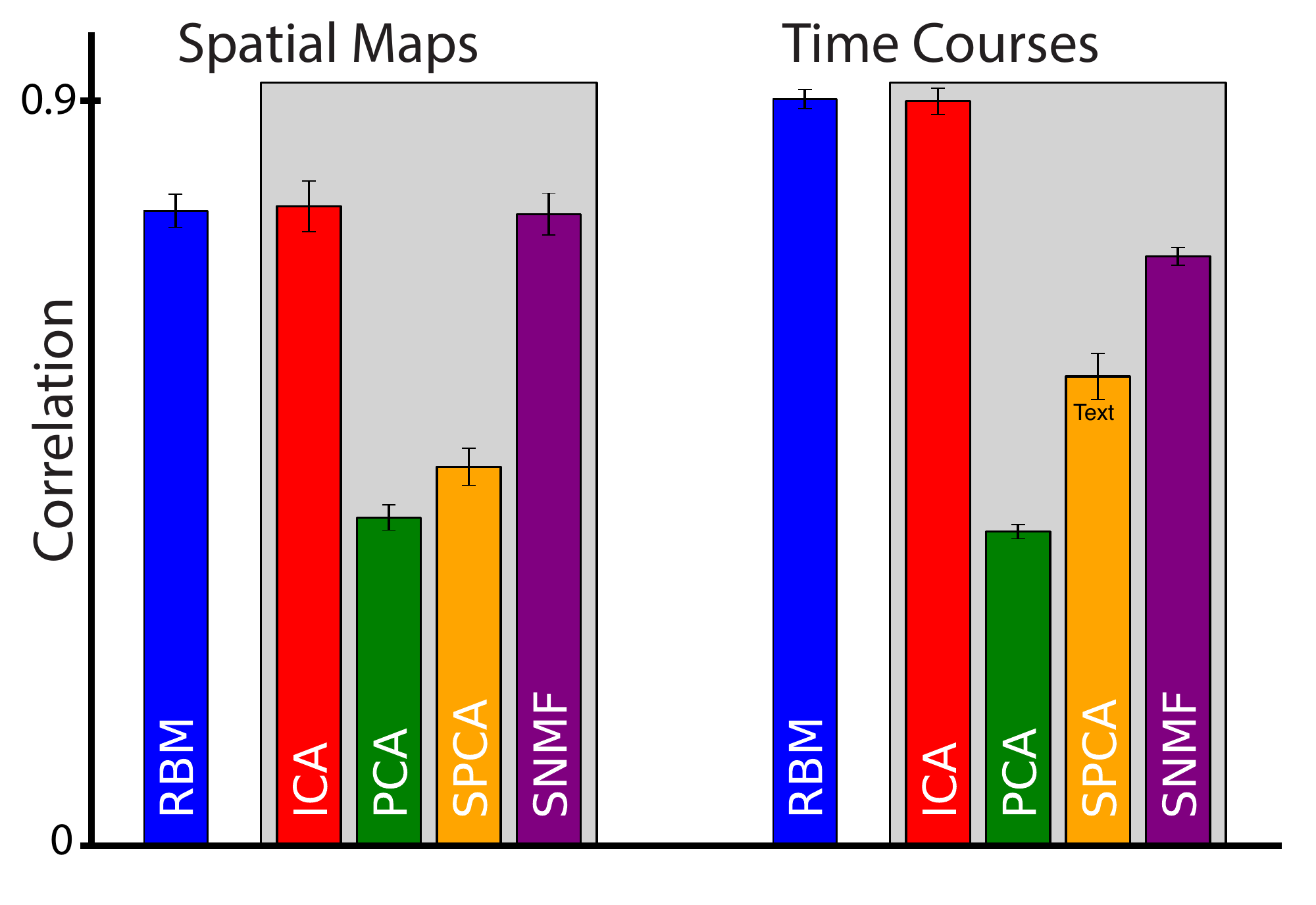}
  \end{minipage}
  \label{fig:other_methods}
}\subfloat[Ground truth (GT) SMs and estimates obtained by RBM and ICA
(thresholded  at \(0.4\)  height).  Colors  are consistent  across the
methods.   Grey  indicates  background  or  areas  without  SMs  above
threshold.]{

  \begin{minipage}[c][0.22\textwidth]{%
      0.3\textwidth}
    \centering%
    \includegraphics[width=1\textwidth]{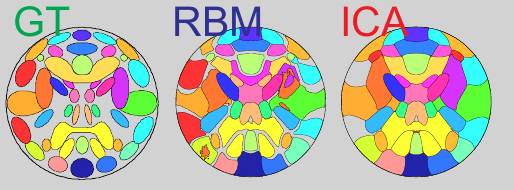}
  \end{minipage}
  \label{fig:synthetic_data}
}%
\subfloat[Spatial, temporal, and cross  correlation (FNC) accuracy for
ICA (red) and RBM (blue), as a function of spatial overlap of the true
sources  from~\ref{fig:synthetic_data}.   Lines indicate  the  average
correlation to GT, and the color-fill indicates $\pm2$ standard errors
around the mean.]{%
  \begin{minipage}[c][0.22\textwidth]{%
      0.36\textwidth}
    \centering%
    \includegraphics[width=1\textwidth]{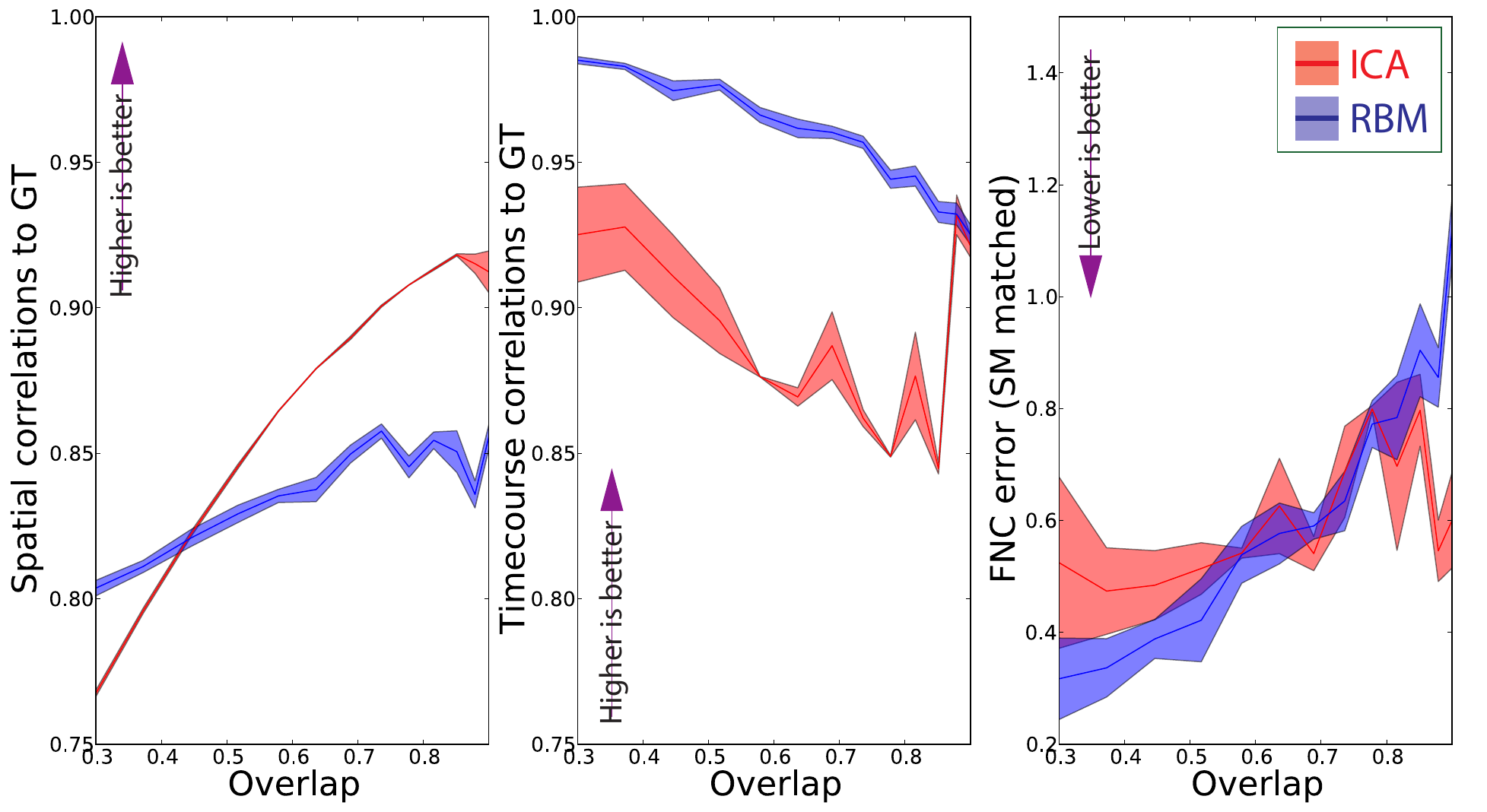}
  \end{minipage}
  \label{fig:overlap_error}
}%
\caption{Comparison of RBM estimation accuracy of features and their
  time courses with SMFs.}
\label{fig:synth}
\end{figure}
In  this  section  we  summarize  our  comparisons  of  RBM  with  SMF
models---including         Infomax        ICA~\citep{BellSejnowski95},
PCA~\citep{trevor2009elements},               sparse               PCA
(sPCA)~\citep{zou2006sparse},          and         sparse          NMF
(sNMF)~\citep{Hoyer2002}---on synthetic  data with known  spatial maps
generated to simulate fMRI.

Figure~\ref{fig:other_methods} shows  the correlation of  spatial maps
(SM) and time course (TC) estimates  to the ground truth for RBM, ICA,
PCA, sPCA, and sNMF.  Correlations are averaged across all sources and
datasets. RBM and ICA showed the best overall performance.  While sNMF
also  estimated  SMs  well,  it  showed  inferior  performance  on  TC
estimation,  likely due  to the  non-negativity constraint.   Based on
these results and the broad adoption of  ICA in the field, we focus on
comparing Infomax ICA and RBM.

Figure~\ref{fig:synthetic_data}  shows the  full set  of ground  truth
sources along with  RBM and ICA estimates for  a single representative
dataset.   SMs  are  thresholded   and  represented  as  contours  for
visualization.   Results over  all synthetic  datasets showed  similar
performance for  RBM and ICA (Figure~\ref{fig:overlap_error}),  with a
slight advantage  for ICA with regard  to SM estimation, and  a slight
advantage for  RBM with regards  to TC  estimation.  RBM and  ICA also
showed  comparable  performance  estimating  cross  correlations  also
called functional network connectivity (FNC).

\subsection{An fMRI data application}
\label{sec:realdata}
\begin{figure}[ht!]
  \centering
  \includegraphics[width=\linewidth]{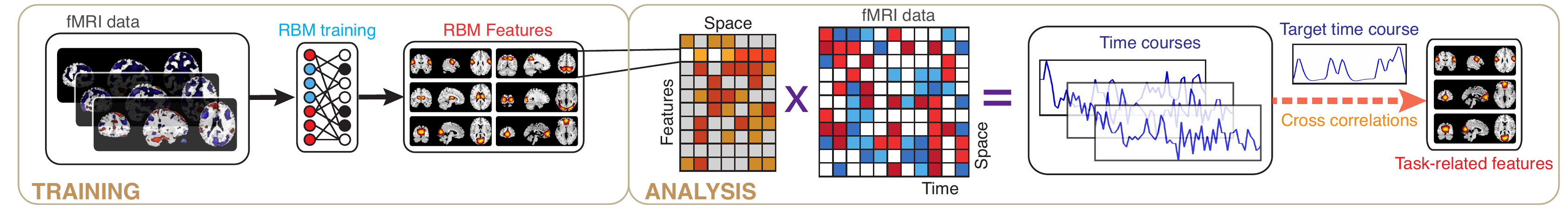}
  \vspace{-10pt}
  \caption{The processes of feature learning and time course
    computation from fMRI data by an RBM. The visible units are voxels
  and a hidden unit receptive field covers an fMRI volume.}
  \vspace{-0pt}
  \label{fig:pipeline}
\end{figure}
Data used in  this work comprised of task-related scans  from 28 (five
females)  healthy participants,  all of  whom gave  written, informed,
IRB-approved  consent at  Hartford Hospital  and were  compensated for
participation\footnote{More detailed information regarding participant
  demographics is provided  in~\cite{swanson2010}}.  All participants
were  scanned during  an  auditory oddball  task  (AOD) involving  the
detection of  an infrequent target  sound within a series  of standard
and  novel  sounds\footnote{The  task  is  described  in  more  detail
  in~\cite{calhoun2008modulation} and~\cite{swanson2010}.}.

Scans were acquired at the Olin Neuropsychiatry Research Center at the
Institute  of  Living/Hartford  Hospital   on  a  Siemens  Allegra  3T
dedicated  head scanner  equipped with  \unitfrac[40]{mT}{m} gradients
and           a           standard           quadrature           head
coil~\cite{calhoun2008modulation,swanson2010}.   The AOD  consisted of
two  8-min runs,  and $249$  scans (volumes)  at 2  second TR  (0.5 Hz
sampling  rate)   were  used  for   the  final  dataset.    Data  were
post-processed         using         the         SPM5         software
package~\cite{friston1994statistical},    motion    corrected    using
INRIalign~\cite{freire2002best}, and subsampled to $\unit[53 \times 63
\times 46] {voxels}$.  The complete fMRI dataset was masked below mean
and the mean  image across the dataset was removed,  giving a complete
dataset of size $70969$ voxels by $6972$ volumes.  Each voxel was then
normalized to have zero mean and unit variance.

\begin{wrapfigure}[12]{R}[0cm]{5cm}
  \vspace{-10pt}
  \centering
  \includegraphics[width=\linewidth]{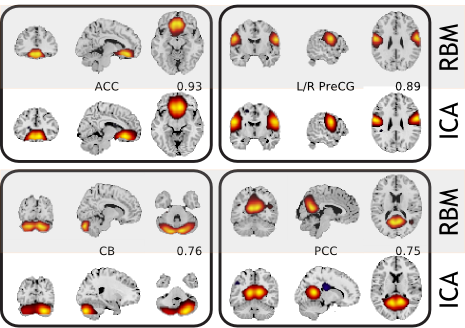}
  \vspace{-15pt}
  \caption{Intrinsic brain networks estimated by ICA and RBM.}
  \vspace{-0pt}
  \label{fig:RBMvsICA}
\end{wrapfigure}
The RBM was constructed using  $70969$ Gaussian visible units and $64$
hyperbolic  tangent hidden  units.   The  hyper parameters  $\epsilon$
(0.08 from the searched $[\num{1e-4}, \num{1e-1}]$ range) for learning
rate    and     $\lambda$    (0.1    from    the     searched    range
[\num{0.01},\num{0.1}]) for $L_1$ weight  decay were selected as those
that showed  a reduction of  reconstruction error over training  and a
significant   reduction    in   span    of   the    receptive   fields
respectively. Parameter  value outside  the ranges either  resulted in
unstable  or slow  learning ($\epsilon$)  or uninterpretable  features
($\lambda$).  The  RBM was then trained  with a batch size  of $5$ for
approximately  $100$  epochs to  allow  for  full convergence  of  the
parameters.

After  flipping  the  sign  of  negative  receptive  fields,  we  then
identified and labeled spatially distinct features as corresponding to
brain  regions  with  the  aid  of  AFNI~\cite{cox1996afni}  excluding
features  which  had a  high  probability  of corresponding  to  white
matter, ventricles, or artifacts (eg. motion, edges).

We normalized  the fMRI volume time  series to mean zero  and used the
trained RBM in feed-forward mode to  compute time series for each fMRI
feature.  This  was done to better  compare to ICA, where  the mean is
removed in PCA preprocessing.

The   work-flow  is   outlined  in   Figure~\ref{fig:pipeline},  while
Figure~\ref{fig:RBMvsICA} shows comparison  of resulting features with
those obtained by Infomax ICA.  In general, RBM performs competitively
with ICA, while  providing--perhaps, not surprisingly due  to the used
$L_1$ regularization---sharper and more  localized features.  While we
recognize that this  is a subjective measure we list  more features in
Figure~\ref{fig:10matches}  of  Section~\ref{sec:supplement} and  note
that  RBM features  lack  negative parts  for corresponding  features.
Note, that  in the  case of $L_1$  regularized weights  RBM algorithms
starts to resemble some of the ICA approaches (such as the recent RICA
by  Le  at  al.~\cite{le2011ica}),   which  may  explain  the  similar
performance. However, the differences  and possible advantages are the
generative  nature  of  the  RBM   and  no  enforcement  of  component
orthogonality  (not  explicit  at  the  least).  Moreover,  the  block
structure  of  the correlation  matrix  (see  below the  Supplementary
material section) of  feature time courses provide a  grouping that is
more  physiologically  supported  than  that  provided  by  ICA.   For
example,  see  Figure~\ref{fig:t_fnc}  in the  supplementary  material
section below.  Perhaps,  because ICA working hard  to enforce spatial
independence   subtly    affects   the   time   courses    and   their
cross-correlations in turn.  We have observed comparable running times
of the (non GPU)  ICA (\url{http://www.nitrc.org/projects/gift}) and a
GPU           implementation            of           the           RBM
(\url{https://github.com/nitishsrivastava/deepnet}).

\vspace{-8pt}
\section{Validating the depth effect}
\label{sec:structural}
Since  the  RBM  results demonstrate  a  feature-learning  performance
competitive  with the  state of  the art  (or better),  we proceed  to
investigating the effects of the model  depth. To do that we turn from
fMRI to  sMRI data.  As  it is commonly  assumed in the  deep learning
literature~\cite{le2010deep}    the   depth    is   often    improving
classification accuracy. We investigate if  that is indeed true in the
sMRI  case. Structural  data is  convenient  for the  purpose as  each
subject/session  is represented  only by  a single  volume that  has a
label:  control or  patient  in our  case. Compare  to  4D data  where
hundreds of volumes  belong to the same subject with  the same disease
state.

\subsection{A deep belief network}
\label{sec:DBN}
A  DBN  is  a  sigmoidal belief  network  (although  other  activation
functions may be used) with an RBM  as the top level prior.  The joint
probability  distribution   of  its   visible  and  hidden   units  is
parametrized as follows:
\begin{align}
  P(\vec{v},\vec{h}^1,\vec{h}^2,\dots,\vec{h}^l)                    &=
  P(\vec{v}|\vec{h}^1)P(\vec{h}^1|\vec{h}^2)\cdots
  P(\vec{h}^{l-2},\vec{h}^{l-1})P(\vec{h}^{l-1},\vec{h}^l),
\end{align}
where $l$ is the number of hidden layers, $P(\vec{h}^{l-1},\vec{h}^l)$
is  an RBM,  and $P(\vec{h}^i|\vec{h}^{i+1})$  factor into  individual
conditionals:
\begin{align}
  P(\vec{h}^i|\vec{h}^{i+1}) &= \prod_{j=1}^{n_i} P(h_j^i|\vec{h}^{i+1})
\end{align}
The important  property of DBN  for our  goals of feature  learning to
facilitate discovery is its ability to operate in generative mode with
fixed values on  chosen hidden units thus allowing  one to investigate
the features that the model have learned and/or weighs as important in
discriminative decisions. We, however, not  going to use this property
in  this section,  focusing instead  on  validating the  claim that  a
network's depth provides benefits for neuroimaging data analysis.  And
we will  do this using  discriminative mode  of DBN's operation  as it
provides an objective measure of the depth effect.

DBN training  splits into two stages:  pre-training and discriminative
fine tuning. A  DBN can be pre-trained by treating  each of its layers
as an RBM---trained in an unsupervised way on inputs from the previous
layer---and later fine-tuned  by treating it as  a feed-forward neural
network.  The latter  allows supervised  training via  the error  back
propagation  algorithm.   We  use  this schema  in  the  following  by
augmenting each DBN with a soft-max layer at the fine-tuning stage.

\subsection{Nonlinear embedding as a constraint satisfaction problem}
\label{sec:dandc}
A DBN and an  RBM operate on data samples, which  are brain volumes in
the fMRI and sMRI case.  A  five-minute fMRI experiment with 2 seconds
sampling  rate yields  150  of  these volumes  per  subject. For  sMRI
studies number of  participating subjects varies but in  this paper we
operate   with   a   300   and  a   3500   subject-volumes   datasets.
Transformations learned by deep learning methods do not look intuitive
in the hidden node space and  generative sampling of the trained model
does not  provide a sense if  a model have learned  anything useful in
the case  of MRI data:  in contrast to  natural images, fMRI  and sMRI
images  do not  look  very  intuitive.  Instead,  we  use a  nonlinear
embedding method to control whether a model learned useful information
and to assist in investigation of what have it, in fact, learned.

One of the purposes of an embedding is to display a complex high
dimensional dataset in a way that is 
\begin{inparaenum}[\itshape   i\upshape)]  
\item intuitive, and
\item representative of the data sample.
\end{inparaenum}
The  first requirement  usually leads  to displaying  data samples  as
points in  a 2-dimensional map, while  the second is more  elusive and
each approach addresses it  differently.  Embedding approaches include
relatively simple random linear projections---provably preserving some
neighbor relations~\cite{deVries:icdm10}---and a more complex class of
nonlinear        embedding        approaches~\cite{van2008visualizing,
  sammon1969nonlinear, tenenbaum2000global,  roweis2000nonlinear}.  In
an attempt to  organize the properties of this diverse  family we have
aimed  at  representing nonlinear  embedding  methods  under a  single
constraint  satisfaction  problem  (CSP) framework  (see  below).   We
hypothesize that each method places the  samples in a map to satisfy a
specific set of  constraints. Although this work is  not yet complete,
it proven useful in our current study. We briefly outline the ideas in
this section to provide enough intuition of the method that we further
use in Section~\ref{sec:structural}.

Since we  can control the constraints  in the CSP framework,  to study
the effect of deep  learning we choose them to do  the least amount of
work---while still being useful---letting the DBN do (or not) the hard
part.      A      more       complicated      method      such      as
t-SNE~\cite{van2008visualizing}  already  does complex  processing  to
preserve the structure of a dataset in a 2D map -- it is hard to infer
if the quality of  the map is determined by a  deep learning method or
the embedding. While some of the existing method may have provided the
``least amount of work'' solutions as well we chose to go with the CSP
framework.   It  explicitly  states  the constraints  that  are  being
satisfied and thus  lets us reason about deep  learning effects within
the constraints, while with  other methods---where the constraints are
implicit---this would have been harder.

A constraint  satisfaction problem (CSP)  is one requiring  a solution
that satisfies a set of constraints. One of the well known examples is
the boolean  satisfiability problem  (SAT).  There are  multiple other
important  CSPs such  as  the packing,  molecular conformations,  and,
recently,    error   correcting    codes~\cite{derbinsky2013improved}.
Freedom to setup  per point constraints without  controlling for their
global   interactions   makes   a  CSP   formulation   an   attractive
representation  of the  nonlinear  embedding  problem.  Pursuing  this
property   we   use  the   iterative   ``divide   and  concur''   (DC)
algorithm~\cite{gravel2008divide}    as    the    solver    for    our
representation.  In DC  algorithm we treat each point  on the solution
map as a  variable and assign a set of  constraints that this variable
needs  to satisfy  (more  on these  later). Then  each  points gets  a
``replica''  for  each  constraint  it  is  involved  into.   Then  DC
algorithm alternates  the divide  and concur projections.   The divide
projection moves each  ``replica'' points to the  nearest locations in
the  2D map  that satisfy  the  constraint they  participate in.   The
concur projection concurs locations of  all ``replicas'' of a point by
placing them at  the average location on  the map. The key  idea is to
avoid local traps by combining the  divide and concur steps within the
difference map~\cite{elser2007searching}. A  single location update is
represented by:
\begin{align}\nonumber
  x_c &= P_{c}((1 + 1/\beta) * P_{d}(x) - 1/\beta * x)\\\nonumber
  x_d &= P_{d}((1 - 1/\beta) * P_{c}(x) + 1/\beta * x)\\
  x &= x + \beta*(x_{c}-x_{d}),
  \label{eq:DM}
\end{align}
where $P_{d}(\cdot)$  and $P_{c}(\cdot)$ denote the  divide and concur
projections and $\beta$ is a user-defined parameter.

While  the   concur  projection  will   only  differ  by   subsets  of
``replicas'' across  different methods representable in  DC framework,
the divide projection is unique and defines the algorithm behavior. In
this  paper, we  choose a  divide  projection that  keeps $k$  nearest
neighbors  of each  point in  the  higher dimensional  space also  its
neighbors  in  the  2D  map.  This  is  a  simple  local  neighborhood
constraint  that  allows  us  to   assess  effects  of  deep  learning
transformation  leaving most  of  the mapping  decisions  to the  deep
learning.

Note, that for  a general dataset we  may not be able  to satisfy this
constraint: each point has exactly the  same neighbors in 2D as in the
original space (and this is what we indeed observe). The DC algorithm,
however, is  only guaranteed  to find  the solution  if it  exists and
oscillates otherwise.   Oscillating behavior is detectable  and may be
used to stop  the algorithm. We found informative watching  the 2D map
in dynamics,  as the points  that keep oscillating  provide additional
information  into  the structure  of  the  data.  Another  practically
important feature  of the algorithm:  it is deterministic.   Given the
same  parameters ($\beta$  and  the parameters  of $P_{d}(\cdot)$)  it
converges to  the same  solution regardless of  the initial  point. If
each of the points participates  in each constraint then complexity of
the  algorithm   is  quadratic.  With  our   simple  $k$  neighborhood
constraints it is $O(kn)$, for $n$ samples/points.

\subsection{A schizophrenia structural MRI dataset}
\label{sec:depth}
\begin{wrapfigure}[14]{R}[0cm]{5.5cm}
  \vspace{-15pt}
  \centering
  \includegraphics[width=5.5cm]{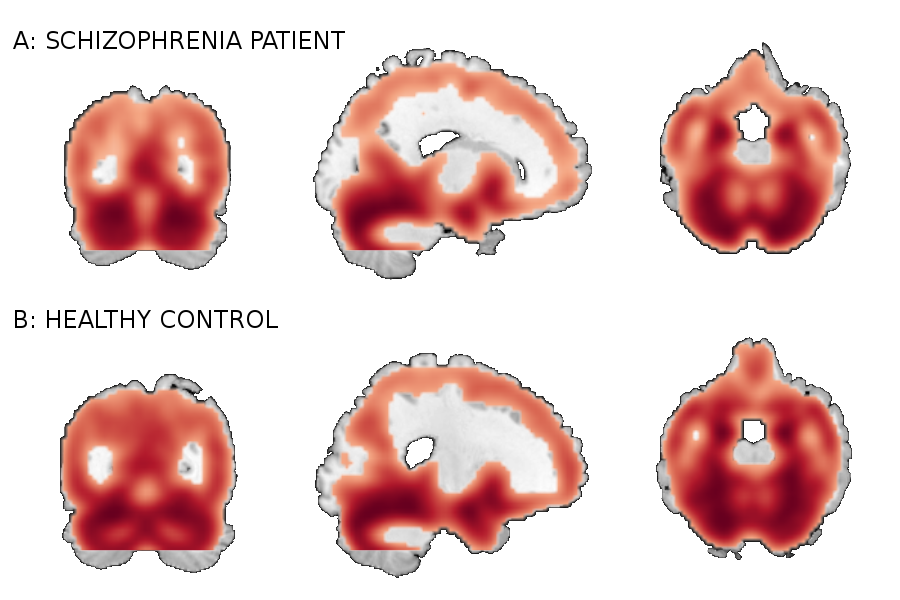}
  \vspace{-15pt}
  \caption{A  smoothed gray  matter segmentation  of a  patient and  a
    healthy control: each is a training sample.}
  \label{fig:smoothGM}
\end{wrapfigure}
We  use  a combined  data  from  four separate  schizophrenia  studies
conducted at Johns Hopkins  University (JHU), the Maryland Psychiatric
Research Center (MPRC), the Institute of Psychiatry, London, UK (IOP),
and the Western Psychiatric Institute  and Clinic at the University of
Pittsburgh (WPIC) (the data used in Meda et al.~\cite{meda2008large}).
The  combined  sample comprised  198  schizophrenia  patients and  191
matched healthy controls and contained  both first episode and chronic
patients~\cite{meda2008large}.  At all  sites, whole  brain MRIs  were
obtained on  a 1.5T  Signa GE scanner  using identical  parameters and
software.   Original structural  MRI images  were segmented  in native
space and  the resulting gray  and white matter images  then spatially
normalized to gray  and white matter templates  respectively to derive
the  optimized normalization  parameters. These  parameters were  then
applied to the whole brain structural  images in native space prior to
a new segmentation.  The obtained  60465 voxel gray matter images were
used in this study. Figure~\ref{fig:smoothGM} shows example orthogonal
slice views of the gray matter data samples of a patient and a healthy
control.

The main  question of this  Section is to  evaluate the effect  of the
depth of  a DBN on sMRI.   To answer this question,  we investigate if
classification rates improve with the  depth. For that we sequentially
investigate DBNs of 3 depth. From RBM experiments we have learned that
even with a larger number of hidden  units (72, 128 and 512) RBM tends
to  only   keep  around  50   features  driving  the  rest   to  zero.
Classification rate and reconstruction  error still slightly improves,
however, when the number of hidden units increases. These observations
affected our choice of 50 hidden units of the first two layers and 100
for the  third.  Each  hidden unit  is connected to  all units  in the
previous layer which  results in an all to  all connectivity structure
between the layers,  which is a more common  and conventional approach
to constructing these models. Note,  larger networks (up to double the
umber of units) lead to similar  results.  We pre-train each layer via
an unsupervised RBM  and discriminatively fine-tune models  of depth 1
(50 hidden units in the top layer), 2 (50-50 hidden units in the first
and the top layer respectively), and  3 (50-50-100 hidden units in the
first,  second and  the top  layer respectively)  by adding  a softmax
layer  on top  of  each of  these  models and  training  via the  back
propagation.

We  estimate   the  accuracy  of  classification   via  10-fold  cross
validation on fine-tuned models splitting the 389 subject dataset into
10 approximately class-balanced folds.
\begin{wraptable}[5]{R}[0cm]{9.5cm}
  \scriptsize
  \vspace{-15pt}
  \begin{center}
    \begin{tabular}{r||cccc}
      depth & raw & 1 & 2 & 3 \\
      \emph{SVM F-score} & \small $0.68\pm0.01$ & \small $0.66\pm0.09$ & \small
      $0.62\pm0.12$ & \small $0.90\pm0.14$ \\
      \emph{LR F-score} & \small $0.63\pm0.09$ & \small $0.65\pm0.11$ & \small
      $0.61\pm0.12$ & \small $0.91\pm0.14$\\
      \emph{KNN F-score} & \small $0.61\pm0.11$ & \small $0.55\pm0.15$ & \small
      $0.58\pm0.16$ & \small $0.90\pm0.16$
    \end{tabular}    
  \end{center}
  \vspace{-12pt}
  \caption{Classification on fine-tuned models (test data)}
  \label{tbl:accuracy}
\end{wraptable}
We  train the  rbf-kernel  SVM, logistic  regression  and a  k-nearest
neighbors (knn) classifier  using activations of the  top-most hidden layers
in  fine-tuned models  to  the training  data of  each  fold as  their
input. The  testing is performed  likewise but  on the test  data.  We
also  perform the  same  10-fold  cross validation  on  the raw  data.
Table~\ref{tbl:accuracy} summarizes the precision and recall values in
the F-scores and their standard deviations.

All models demonstrate a similar trend when the accuracy only slightly
increases from depth-1 to depth-2 DBN and then improves significantly.
Table~\ref{tbl:accuracy} supports  the general claim of  deep learning
community about improvement of classification rate with the depth even
for sMRI data.  Improvement in  classification even for the simple knn
classifier indicates the character of  the transformation that the DBN
learns and applies  to the data: it may be  changing the data manifold
to  organize  classes  by  neighborhoods.  Ideally,  to  make  general
conclusion  about  this  transformation  we need  to  analyze  several
representative datasets.  However, even working  with the same data we
can have a closer view of the depth effect using the method introduced
in Section~\ref{sec:dandc}.
\begin{figure}[ht!]
  \vspace{-5pt}
  \centering
  \includegraphics[width=\linewidth]{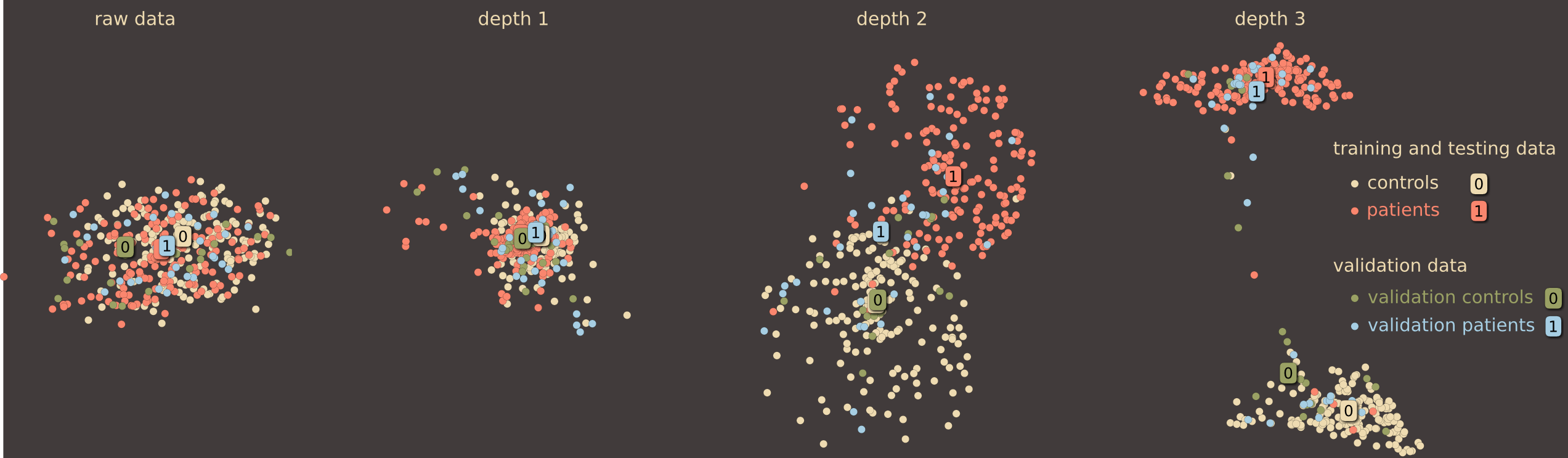}
  \vspace{-0pt}
  \caption{Effect of  a DBN's  depth on neighborhood  relations.  Each
    map is shown at the same  iteration of the algorithm with the same
    $k=50$.   The  color  differentiates  the  classes  (patients  and
    controls)  and the  training  (335 subjects)  from validation  (54
    subjects) data. Although the data becomes separable at depth 1 and
    more so at depth 2, the DBN continues distilling details that pull
    the classes further apart.  }
  \label{fig:changes}
  \vspace{-10pt}
\end{figure}
\vspace{-5pt} 
Although  it  may seem  that  the  DBN  does not  provide  significant
improvements in  sMRI classification from  depth-1 to depth-2  in this
model, it keeps  on learning potentially useful  transformaions of the
data.   We can  see  that using  our  simple local  neighborhood-based
embedding.  Figure~\ref{fig:changes} displays 2D maps of the raw data,
as well as the depth 1, 2,  and 3 activations (of a network trained on
335 subjects): the  deeper networks place patients  and control groups
further apart.  Additionally, Figure~\ref{fig:changes} displays the 54
subjects that the  DBN was not train on.  These  hold out subjects are
also getting increased separation with  depth.  This DBN's behavior is
potentially useful  for generalization,  when larger and  more diverse
data become available.

Our new mapping method has  two essential properties to facilitate the
conclusion and provide confidence in the result: its already mentioned
local properties and  the deterministic nature of  the algorithm.  The
latter leads to  independence of the resulting maps  from the starting
point.  The map only depends on the models parameter $k$---the size of
the neighborhood---and the data.
\begin{wrapfigure}[13]{R}[0cm]{5.5cm}
  \vspace{-0pt}
  \centering
  \includegraphics[width=5.5cm]{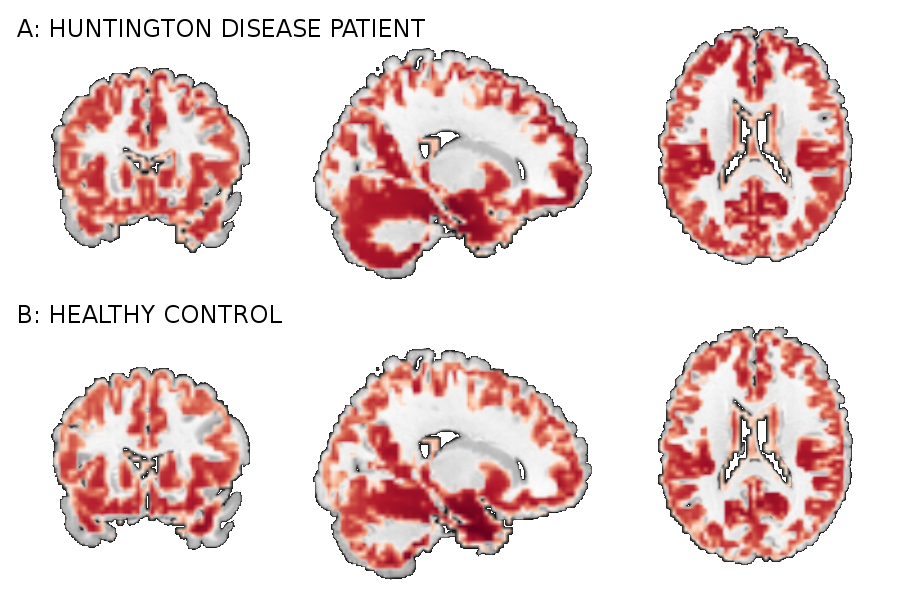}
  \vspace{-15pt}
  \caption{A gray matter  of MRI scans of an HD  patient and a healthy
    control.}
  \label{fig:sharpGM}
\end{wrapfigure}

\subsection{A large-scale Huntington disease data}
\label{sec:discovery}
In this section we focus on  sMRI data collected from healthy controls
and Huntington disease (HD) patients as part of the PREDICT-HD project
(\url{www.predict-hd.net}).   Huntington   disease    is   a   genetic
neurodegenerative disease  that results in degeneration  of neurons in
certain areas of the brain.  The project is focused on identifying the
earliest  detectable changes  in thinking  skills, emotions  and brain
structure  as a  person begins  the  transition from  health to  being
diagnosed  with Huntington  disease. We  would  like to  know if  deep
learning methods can assist in answering that question.

For this study T1-weighted scans  were collected at multiple sites (32
international sites), representing multiple  field strengths (1.5T and
3.0T) and multiple manufactures (Siemens, Phillips, and GE).  The 1.5T
T1 weighted  scans were an  axial 3D volumetric  spoiled-gradient echo
series  ($\approx1\times1\times1.5$  mm  voxels),   and  the  3.0T  T1
weighted  scans   were  a   3D  Volumetric  MPRAGE   series  ($\approx
1\times1\times1$ mm voxels).
\begin{wraptable}[4]{R}[0cm]{9.5cm}
  \scriptsize
  \vspace{-15pt}
  \begin{center}
    \begin{tabular}{r||cccc}
      depth & raw & 1 & 2 & 3 \\
      \emph{SVM F-score} & \small $0.75$ & \small $0.65\pm0.01$ & \small
      $0.65\pm0.01$ & \small $1.00\pm0.00$ \\
      \emph{LR F-score} & \small $0.79$ & \small $0.65\pm0.01$ & \small
      $0.65\pm0.01$ & \small $1.00\pm0.00$\\
    \end{tabular}    
  \end{center}
  \vspace{-10pt}
  \caption{Classification on fine-tuned models (HD data)}
  \label{tbl:Iaccuracy}
\end{wraptable}

The images were segmented in the  native space and the normalized to a
common  template.   After  correlating   the  normalized  gray  matter
segmentation  with the  template  and  eliminating poorly  correlating
scans we obtain a dataset of 3500 scans, where 2641 were from patients
and 859 from healthy controls.
We have used  all of the scans in this  imbalanced sample to pre-train
and  fine  tune   the  same  model  architecture   (50-50-100)  as  in
Section~\ref{sec:depth}  for all  three depths\footnote{Note,  in both
  cases we have  experimented with larger layer sizes  but the results
  were not significantly different  to warrant increase in computation
  and parameters needed to be estimated.}. 
\begin{wrapfigure}[25]{R}[0cm]{6cm}
  \vspace{-15pt}
  \centering
  \includegraphics[width=6cm]{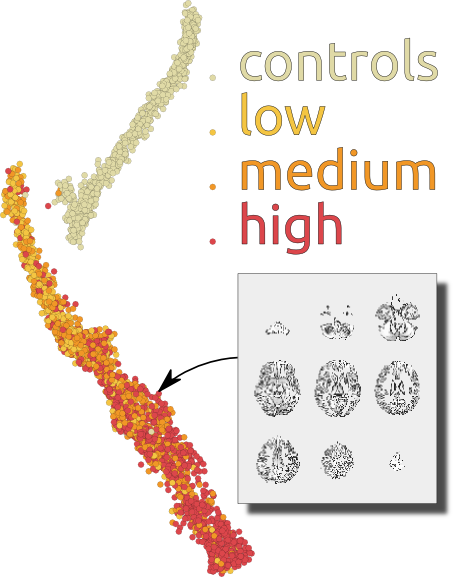}
  \vspace{-15pt}
  \caption{Patients and controls group  separation map with additional
    unsupervised  spectral  decomposition  of sMRI  scans  by  disease
    severity. The map represents 3500 scans.}
  \label{fig:severity}
\end{wrapfigure}

Table~\ref{tbl:Iaccuracy} lists  the average  F-score values  for both
classes at the raw  data and all depth levels. Note  the drop from the
raw data  and then  a recovery  at depth 3.   The limited  capacity of
levels 1  and 2 has reduced  the network ability to  differentiate the
groups but  representational capacity  of depth 3  network compensates
for the initial bottleneck. This, confirms our previous observation on
the depth effect, however, does not  yet help the main question of the
PREDICT-HD  study.  Note,  however, while  Table~\ref{tbl:accuracy} in
the  previous section  evaluates  generalization ability  of the  DBN,
Table~\ref{tbl:Iaccuracy}  here  only  demonstrates changes  in  DBN's
representational capacity  with the depth  as we use no  testing data.
To  further investigate  utility  of the  deep  learning approach  for
scientific discovery we again augment  it with the embedding method of
Section~\ref{sec:dandc}.  Figure~\ref{fig:severity}  shows the  map of
3500 scans of HD patients and healthy controls.  Each point on the map
is    an    sMRI    volume,   shown    in    Figures~\ref{fig:sharpGM}
and~\ref{fig:severity}.  Although  we have  used the complete  data to
train the  DBN, discriminative fine-tuning  had access only  to binary
label: control  or patient. In  addition to that, we  have information
about severity of  the disease from low to high.   We have color coded
this information in Figure~\ref{fig:severity} from bright yellow (low)
through orange (medium) to  red (high). The network\footnote{Note, the
  embedding algorithm does not have  access to any label information.}
discriminates  the patients  by disease  severity which  results in  a
spectrum on the map. Note, that neither t-SNE (not shown), nor our new
embedding  see the  spectrum or  even the  patient groups  in the  raw
data. This is a important property of the method that may help support
its  future use  in discovery  of new  information about  the disease.
\vspace{-10pt}
\section{Conclusions}
\label{sec:conclusions}
\vspace{-10pt} Our investigations  show that deep learning  has a high
potential  in  neuroimaging applications.   Even  the  shallow RBM  is
already competitive  with the  model routinely used  in the  field: it
produces  physiologically meaningful  features  which are  (desirably)
highly focal and have time course cross correlations that connect them
into meaningful functional  groups (Section~\ref{sec:supplement}). The
depth of the  DBN does indeed help classification  and increases group
separation.  This is  apparent on  two sMRI  datasets collected  under
varying  conditions, at  multiple sites  each, from  different disease
groups, and pre-processed  differently.  This is a  strong evidence of
DBNs robustness.   Furthermore, our  study shows  a high  potential of
DBNs   for   exploratory   analysis.    As   Figure~\ref{fig:severity}
demonstrates,  DBN in  conjunction  with our  new  mapping method  can
reveal hidden relations in data. We did find it difficult initially to
find workable  parameter regions, but  we hope that  other researchers
won't have this difficulty starting  from the baseline that we provide
in this paper.


\bibliography{fnc,papers}
\bibliographystyle{plain}
\pagebreak
\section{Supplementary material}
\label{sec:supplement}
\setcounter{figure}{0}
\makeatletter 
\renewcommand{\thefigure}{S\@arabic\c@figure}
\makeatother

\begin{figure*}[!ht]
\includegraphics[width=1\linewidth]{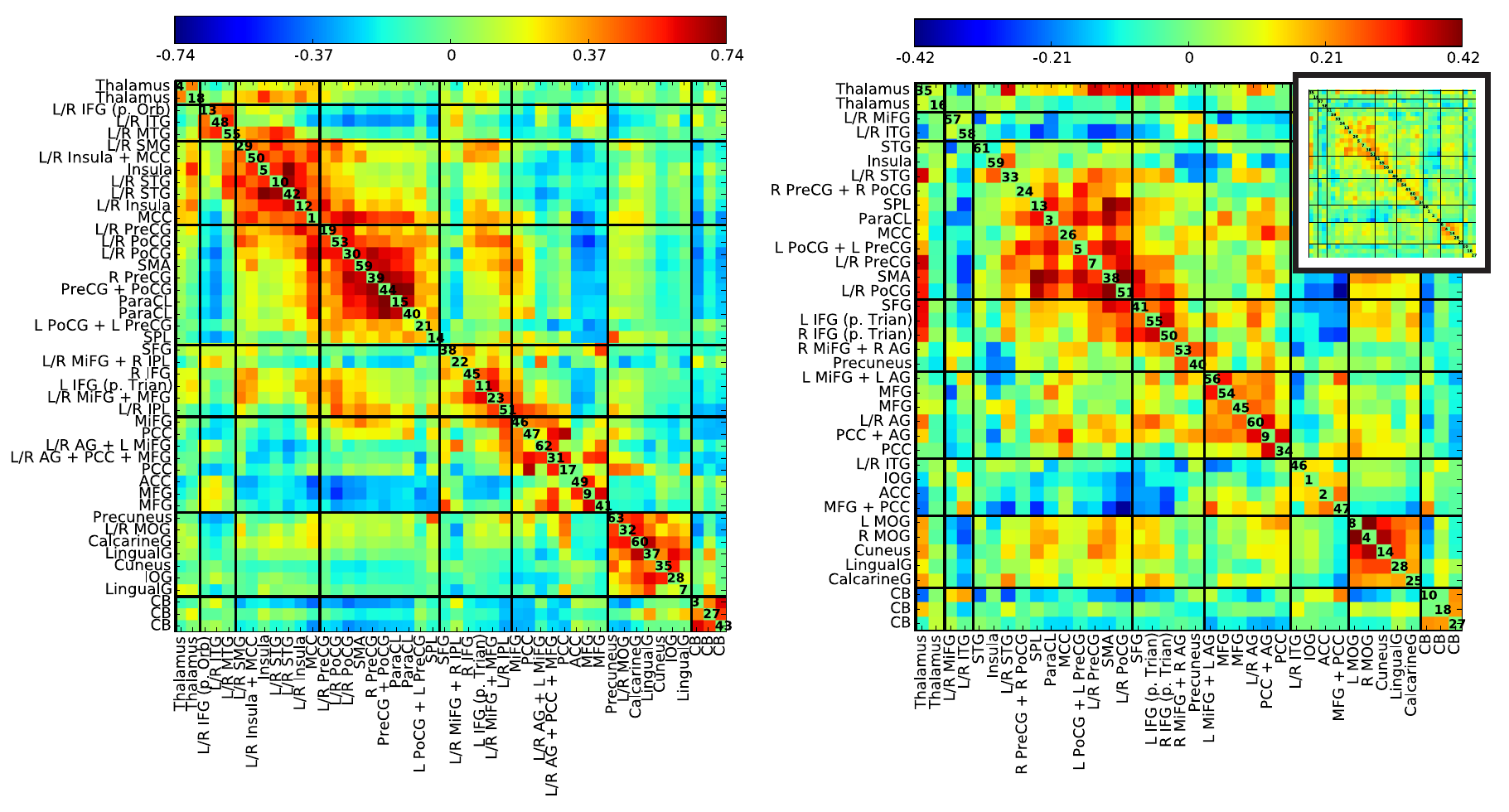}
\caption{Correlation  matrices  determined  from RBM  (left)  and  ICA
  (right), averaged over subjects.  Note that the color scales for RBM
  and ICA  are different (RBM  shows a larger range  in correlations).
  The correlation  matrix for  ICA on  the same scale  as RBM  is also
  provided as  an inset (upper  right). Feature groupings for  RBM and
  ICA  were determined  separately using  the FNC  matrices and  known
  anatomical and functional properties.}
\label{fig:t_fnc}
\end{figure*}

The correlation  matrices for  both RBM  and ICA  results on  the fMRI
dataset    of     Section~\ref{sec:realdata}    are     provided    in
Figure~\ref{fig:t_fnc}, where the ordering  of components is performed
separately  for   each  method.  Each   network  is  named   by  their
physiological function but  we do not go in depth  explaining these in
the  current  paper.   For  RBM, modularity  is  more  apparent,  both
visually     and    quantitatively.      Modularity,    as     defined
in~\citep{rubinov2011}, averages $0.40 \pm  0.060$ across subjects for
RBM, and  $0.35 \pm  0.056$ for ICA.   These values  are significantly
greater for RBM ($t = 7.15, p  < 1e{-6}$ per the paired t-test).  Also
note  that  the  scale  of  correlation values  for  RBM  and  ICA  is
different, which highlights that RBM overestimated strong FNC values.

\begin{figure*}[!ht]
\includegraphics[width=1\linewidth]{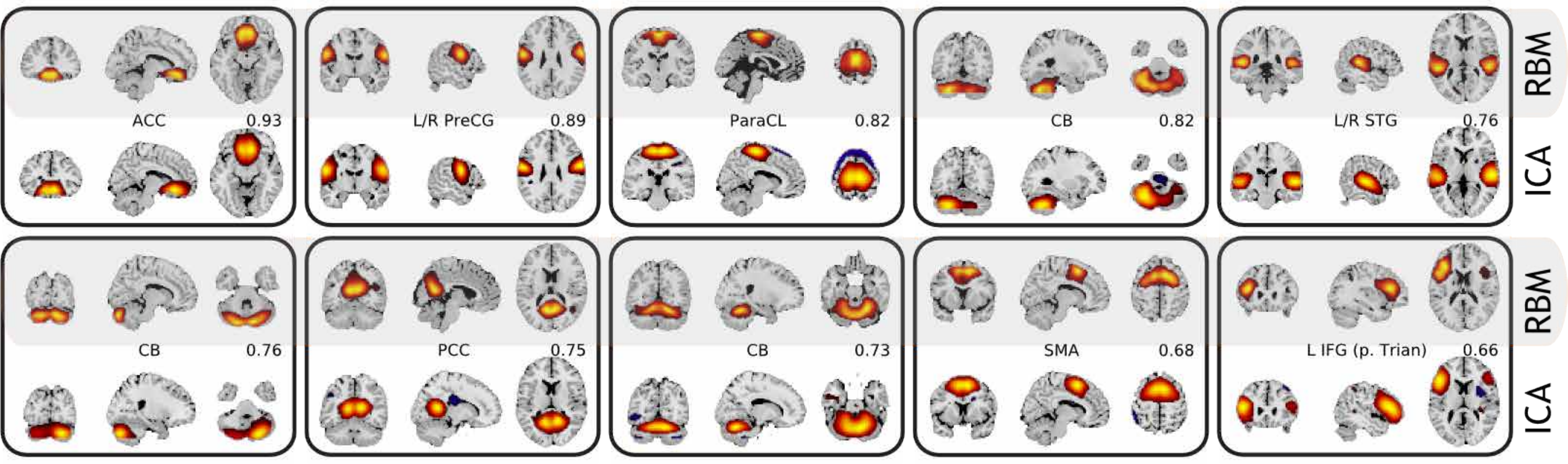}
\caption{Sample pairs  consisting of  RBM (top)  and ICA  (bottom) SMs
  thresholded at $2$  standard deviations.  Pairing was  done with the
  aid  of  spatial  correlations,   temporal  properties,  and  visual
  inspection.  Values indicate the spatial correlation between RBM and
  ICA SMs.}
\label{fig:10matches}
\end{figure*}

\end{document}